\documentclass[twoside]{article}

\usepackage[accepted]{style/aistats2023}

\usepackage{microtype}
\usepackage[utf8]{inputenc}
\usepackage[T1]{fontenc}
\usepackage{latexsym}
\usepackage{amsmath}
\usepackage{amssymb}
\usepackage{amsthm}
\usepackage{relsize}
\usepackage{xspace}
\usepackage[hyphens,spaces,obeyspaces]{url}
\usepackage{enumitem}

\usepackage[group-separator={,}]{siunitx}

\usepackage{tikz}
\usetikzlibrary{bayesnet}
\usetikzlibrary{arrows}
\usepackage{color}
\usepackage{graphicx}
\usepackage{caption}
\usepackage{subcaption}
\usetikzlibrary{backgrounds}

\usepackage{hhline}
\usepackage{xcolor}
\usepackage{colortbl}
\definecolor{grey}{RGB}{150,150,150}

\newcommand{\tikzcircle}[2][red,fill=red]{\tikz[baseline=-0.5ex]\draw[#1,radius=#2] (0,0) circle ;}%

\newtheorem{proposition}{Proposition}[section]

\def\expect{{\mathbb{E}}}

\def\valpha{{\boldsymbol{\alpha}\xspace}}

\def\vpi{{\boldsymbol{\pi}\xspace}}
\def\vphi{{\boldsymbol{\phi}\xspace}}

\newcommand{\defn}[1]{\textit{#1}}

\newcommand{\setquote}[1]{``{#1}''}

\usepackage{algorithm}
\usepackage{algpseudocode}

\usepackage{bbm}
\usepackage{todonotes}
\makeatletter
\newcommand*\iftodonotes{\if@todonotes@disabled\expandafter\@secondoftwo\else\expandafter\@firstoftwo\fi}  
\makeatother

\newcommand{\note}[4][]{\todo[author=#2,color=#3,size=\scriptsize,fancyline,caption={},#1]{#4}} 

\newcommand{\ben}[2][]{\note[#1]{\ben}{green!60}{#2}}

\usepackage{tipa}

\newcommand{\ethz}{\text{\normalfont \textipa{Q}}}
\newcommand{\uchicago}{\normalfont \text{\textipa{@}}}
\newcommand{\unc}{\normalfont \text{\textipa{N}}}

\newcommand{\yijat}{y^{(t)}_{i\xrightarrow{a}j}}
\newcommand{\yat}{y^{(t)}_{a}}

\newcommand{\lambdaijkt}{\lambda^{(t)}_{i\xrightarrow{k}j}}
\newcommand{\lambdakt}{\lambda^{(t)}_{k}}

\newcommand{\tp}{\!+\!}
\newcommand{\tm}{\!-\!}
\newcommand{\teq}{\!=\!}

\newcommand{\bV}{V}
\newcommand{\bA}{A}

\def\tY{{\boldsymbol{Y}\xspace}}
\def\tLambda{\boldsymbol{\Lambda}\xspace}

\newcommand{\OMD}{Ordered Matrix Dirichlet\xspace}
\newcommand{\omd}{\text{OMD}\xspace}

\newcommand{\BMD}{Banded Matrix Dirichlet\xspace}
\newcommand{\bmd}{\text{BMD}\xspace}

\newcommand{\SMD}{Standard Matrix Dirichlet\xspace}
\newcommand{\smd}{\text{SMD}\xspace}

\newcommand{\DPTM}{Dynamic Poisson Tucker model\xspace}
\newcommand{\DPT}{Dynamic Poisson Tucker\xspace}
\newcommand{\dpt}{\text{DPT}\xspace}
\newcommand{\ssm}{SSM\xspace}
\newcommand{\ssms}{SSMs\xspace}

\newcommand{\sppd}{SPPD\xspace}

\newcommand{\link}{https://github.com/niklasstoehr/ordered-matrix-dirichlet}

\definecolor{darkblue}{RGB}{5, 0, 144}
\usepackage[colorlinks,linktoc=all,breaklinks=true]{hyperref}
\usepackage[all]{hypcap}
\hypersetup{citecolor=darkblue}
\hypersetup{linkcolor=darkblue}
\hypersetup{urlcolor=darkblue}

\usepackage{cleveref}
\crefname{section}{\S}{\S\S}
\Crefname{section}{\S}{\S\S}
\crefname{table}{Tab.}{}
\crefname{figure}{Fig.}{}
\crefname{algorithm}{Algorithm}{}
\crefname{equation}{Eq.}{}
\crefname{appendix}{App.}{}
\crefname{thm}{Theorem}{}
\crefname{prop}{Proposition}{}
\crefname{cor}{Corollary}{}
\crefname{observation}{Observation}{}
\crefname{assumption}{Assumption}{}
\crefformat{section}{\S#2#1#3}

\makeatletter
\newcommand\footnoteref[1]{\protected@xdef\@thefnmark{\ref{#1}}\@footnotemark}
\makeatother

\usepackage[round]{natbib}

\begin{document}

\runningtitle{The Ordered Matrix Dirichlet for State-Space Models}

\runningauthor{Niklas Stoehr, Benjamin J. Radford, Ryan Cotterell, Aaron Schein}

\twocolumn[

\aistatstitle{The Ordered Matrix Dirichlet for State-Space Models}

\aistatsauthor{Niklas Stoehr$^{\ethz}$ ~\;~ ~\;~ ~\;~ Benjamin J. Radford$^{\unc}$ ~\;~ ~\;~ ~\;~ Ryan Cotterell$^{\ethz}$ ~\;~ ~\;~ ~\;~ Aaron Schein$^{\uchicago}$}
\aistatsaddress{$^{\ethz}$ETH Zurich ~\;~ ~\;~ ~\;~ $^{\unc}$UNC Charlotte ~\;~ ~\;~ ~\;~ $^{\uchicago}$The University of Chicago\\
\scriptsize{\href{mailto:niklas.stoehr@inf.ethz.ch}{\texttt{niklas.stoehr@inf.ethz.ch}}} ~\;~ ~\;~
\scriptsize{\href{mailto:bradfor7@uncc.edu}{\texttt{bradfor7@uncc.edu}}} ~\;~ ~\;~
\scriptsize{\href{mailto:ryan.cotterell@inf.ethz.ch}{\texttt{ryan.cotterell@inf.ethz.ch}}} ~\;~ ~\;~
\scriptsize{\href{mailto:schein@uchicago.edu}{\texttt{schein@uchicago.edu}}}
 }
]

\begin{abstract}
Many dynamical systems in the real world are naturally described by latent states with intrinsic ordering, such as \setquote{ally}, \setquote{neutral}, and \setquote{enemy} relationships in international relations. These latent states manifest through countries' cooperative versus conflictual interactions over time. \emph{State-space models (\ssms)} explicitly relate the dynamics of observed measurements to transitions in latent states. For discrete data, \ssms commonly do so through a state-to-action \emph{emission matrix} and a state-to-state \emph{transition matrix}. This paper introduces the \defn{\OMD (\omd)} as a prior distribution over ordered stochastic matrices wherein the discrete distribution in the $k^{\textrm{th}}$ row is stochastically dominated by the $(k \tp 1)^{\textrm{th}}$, such that probability mass is shifted to the right when moving down rows. We illustrate the \omd prior within two \ssms: a hidden Markov model, and a novel dynamic Poisson Tucker decomposition model tailored to international relations data. We find that models built on the \omd recover interpretable ordered latent structure without forfeiting predictive performance. We suggest future applications to other domains where models with stochastic matrices are popular (e.g., topic modeling), and publish \href{\link}{user-friendly code}.~\looseness=-1
\end{abstract}



\begin{figure}[t]
\centering
\includegraphics[width=1.0\linewidth]{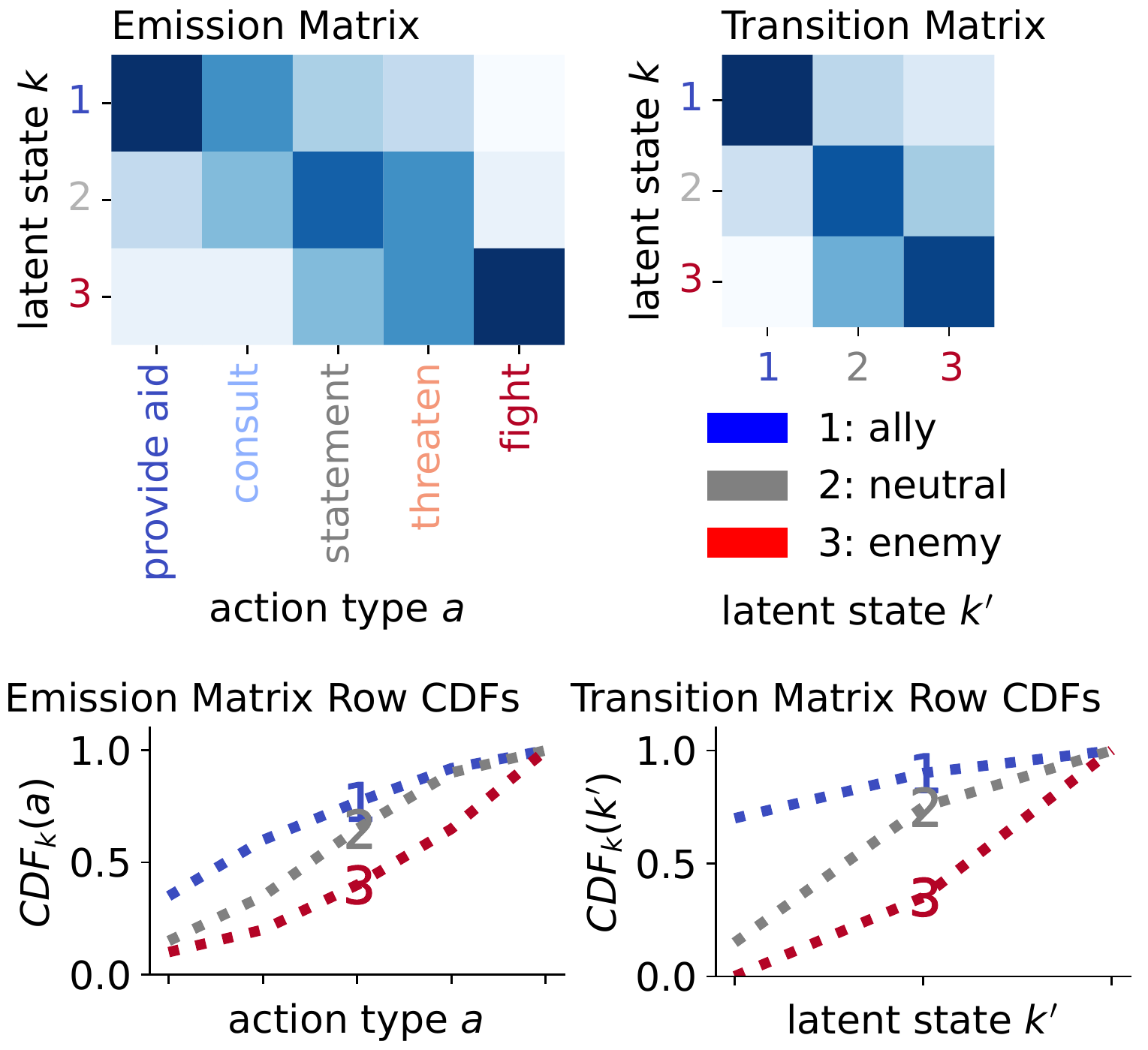}
\caption{Action types in international relations data are ordered along a cooperation-to-conflict axis. Our model infers latent states with an ordering that reflects the ordering in observed actions. \emph{Emission matrix}: more conflictual actions (indexed by higher $a$), are generated by more conflictual latent states (higher $k$). \emph{Transition matrix}: latent states transition to neighboring ones. \emph{CDFs}: The \OMD enforces that the CDF of the $k^{\textrm{th}}$ discrete distribution is always greater than the $(k\tp 1)^{\textrm{th}}$, so that probability mass in the stochastic matrix shifts right when moving down rows.\looseness=-1}       
\label{fig:emission_transition} 
\end{figure}

\section{INTRODUCTION}
\label{sec:intro}

In many modeling settings and application domains, some aspect of the observation space has intrinsic ordering. For example, in international relations, observed interactions between countries can be ordered on a conflict-to-cooperation axis, ranging from \setquote{provide aid} to \setquote{fight} \citep{goldstein_conflict-cooperation_1992, schrodt_kansas_2008}. This ordering should ideally be reflected in any state space used to summarize or describe observed interactions. For example, more conflictual actions like \setquote{fight} or \setquote{threaten} might be more likely between countries in an \setquote{enemy} state than those in an \setquote{ally}state~\citep{schrodt_forecasting_2006}. In this example, the latent states represent relationship statuses between countries, ordered from \setquote{ally} to \setquote{enemy}, and reflect the conflict-to-cooperation ordering of the observed actions. We might expect states to transition to other states over time in a way that also reflects their intrinsic ordering. Allies rarely become enemies from one moment to the next, but rather (de-)escalate gradually, passing first through intermediate states~\citep{rand-escalation-1984}.\looseness=-1

State-space models (\ssms) are statistical models that explicitly relate time-varying measurements to latent states, such that patterns and trends in the observed space are attributable to transitions between states over time. For discrete data, the canonical form of such models is based on two stochastic matrices: the \emph{emission matrix}, which describes how latent states generate observations, and the \emph{transition matrix}, which describes how states transition to other states over time. This general formulation does not intrinsically promote any ordering of the latent states, whose indices are arbitrary and subject to \setquote{label switching}~\citep{richardson_bayesian_1997,stephens_dealing_2000}.

To promote some sense of ordering in the state space, researchers sometimes constrain the transition matrix to take only banded or ``left-right-left'' forms~\citep{schrodt_forecasting_2006,netzer_hidden_2008,randahl_predicting_2022}, whereby states only transition to adjacent states. This constraint substantially restricts the expressiveness of the model while still not ensuring a well-defined ordering of the latent states that reflects the ordering in the observation space.

This paper introduces a novel prior distribution over stochastic matrices, the \defn{\OMD} (\omd), and demonstrates it as a key ingredient in \ssms with well-ordered state spaces. An \omd random variable is a stochastic matrix whose rows are discrete distributions that sum to 1, and whose $k^{\textrm{th}}$ row is stochastically dominated by the $(k\tp 1)^{\textrm{th}}$, so that probability mass shifts to the right when moving down the matrix (\Cref{fig:emission_transition}). We define the \omd distribution implicitly via a stick-breaking construction that ensures the desired ordering property by sorting Beta-distributed auxiliary variables. As we show, when the \omd is selected as a prior over both emission and transition matrices in an \ssm, the inferred latent states have an intrinsic ordering that reflects ordering in the observation space.\looseness=-1

To demonstrate and evaluate the \omd as a prior, we construct and apply two different \ssms---(1) a simple hidden Markov model (HMM) that we apply to synthetic data where the ground-truth latent structure is known, and (2) a novel dynamic version of Bayesian Poisson Tucker decomposition~\citep{schein_bayesian_2016}, which we apply to international relations data of country-to-country interactions. 

We compare each of these models to a baseline that differs only in its prior over the transition and emission matrices---instead of the \omd, it makes the standard assumption of rows being independently Dirichlet-distributed, which we term the \SMD (\smd). We find that the models based on the \omd are more readily interpretable than those based on the \smd while still performing comparably and sometimes better in forecasting and imputation tasks. In synthetic experiments, the \omd model is much more effective at recovering ground-truth latent structure, while on international relations data, the \omd model exhibits superior forecasting performance over both the \smd model and an additional baseline we introduce that constrains the transition matrix to be banded.~\looseness=-1

After setting up notation and providing background on \ssms in~\Cref{sec:ssms}, we formally introduce the \omd in~\cref{sec:omd} and motivate it as a prior within \ssms. In~\Cref{sec:inference} we discuss posterior inference for \omd models using Pyro~\citep{bingham_pyro_2018}. We then provide results from a suite of synthetic data experiments in~\Cref{sec:synthetic_data}, and present a case study on international relations data in~\Cref{sec:application}. Finally, we discuss broader connections in~\Cref{sec:discussion}, and summarize our conclusions in~\Cref{sec:conclusion}.

\section{STATE-SPACE MODELS}
 \label{sec:ssms}
 
State-space models (\ssms) describe the evolution of time-indexed measurements $\boldsymbol{y}^{(t)}$ in terms of corresponding latent states $\boldsymbol{\lambda}^{(t)}$ \citep{kalman_new_1960}. \ssms assume that patterns and trends in the observed measurements, typically only noisily realized, are attributable to transitions between latent states.~\looseness=-1

\paragraph{Basic Form} This paper considers a subset of \ssms frequently used to model discrete or non-negative data. Consider a non-negative vector-valued measurement $\boldsymbol{y}^{(t)} \in \mathbb{R}_+^{\bA}$ at discrete time step $t$. Using the international relations example in the introduction, $\boldsymbol{y}^{(t)}$ might measure the counts of $A$ different action types taken between some pair of countries during time step $t$. We use $a \in [A]$ to index into this vector, so that $\yat$ is an entry, and refer to $a$ as an \emph{action} or \emph{action type} throughout.\looseness=-1

The \ssms we consider connect observed measurements to vector-valued, non-negative latent states $\boldsymbol{\lambda}^{(t)} \in \mathbb{R}_+^{K}$ under the following assumption:
\begin{align}
    \mathbb{E}[\yat] &\propto \sum_{k=1}^{K} \lambdakt\hspace{-0.35em}\underbrace{\phi_{ka}}_\text{\tiny emission}
    \label{eq:emission}
\end{align}
where $\phi_{ka} \in [0,1]$ is an entry in the discrete distribution $\boldsymbol{\phi}_k$ which sums to one over actions $\sum_{a=1}^A \phi_{ka} \teq 1$ and is itself the $k^{\textrm{th}}$ row of the state-to-action \emph{emission matrix} $\Phi$.

The states then evolve over time under the assumption
\begin{align}
    \expect[\lambdakt] &\propto \sum_{k'=1}^K \lambda^{(t\tm 1)}_{k'} \hspace{-0.35em} \underbrace{\pi_{k'k}}_\text{\tiny transition}
    \label{eq:transition}
\end{align}
where $\pi_{k'k} \in [0,1]$ is an entry in the discrete distribution $\boldsymbol{\pi}_{k'}$ which is itself the $(k')^{\textrm{th}}$ row of the state-to-state \emph{transition matrix} $\Pi \in [0,1]^{K \times K}$.

Many \ssms follow this basic form in ~\cref{eq:emission,eq:transition}, such as hidden Markov models (HMMs), discrete dynamical systems~\citep{schein_poisson-gamma_2016}, or more complex \ssms as presented in \cref{sec:dptm}. The key feature of these models is that they involve an emission $\Phi$ and transition matrix $\Pi$ which are both \emph{(row-)stochastic matrices}.

\paragraph{What is a \setquote{State}?} There are differences in the \ssm literature on whether the \setquote{state} at time step $t$ \emph{is} the vector $\boldsymbol{\lambda}^{(t)}$, or whether each element $\lambdakt$ of the vector describes the relevance of one of $k \in [K]$ \setquote{states}. These two interpretations coincide when $\boldsymbol{\lambda}^{(t)}$ is a one-hot vector, placing non-zero mass on only one element, as in HMMs. However, these interpretations diverge in more general settings. We adopt both senses of the word \setquote{state} in this paper, referring to $\boldsymbol{\lambda}^{(t)}$ as the \emph{complex \setquote{state} of the overall system} at $t$ but also understanding it as a \emph{mixture over $K$ simple \setquote{states}}.

\paragraph{Dirichlet Priors} Researchers often place prior distributions over model parameters either as a way to encode structural assumptions about the state space or to fit models using Bayesian inference (or both). The conventional prior for row-stochastic matrices assumes that rows are independently Dirichlet distributed, what we will refer to as the \defn{\SMD (\smd)}. A draw $\boldsymbol{\phi} \sim \textrm{Dir}(\boldsymbol{\alpha})$ from a Dirichlet distribution with concentration parameter $\boldsymbol{\alpha} \in \mathbb{R}_+^A$ is a discrete distribution over $A$ categories, $\sum_{a=1}^A \phi_a \teq 1$.\looseness=-1

\begin{figure}[t]
\centering
\includegraphics[width=1.0\linewidth]{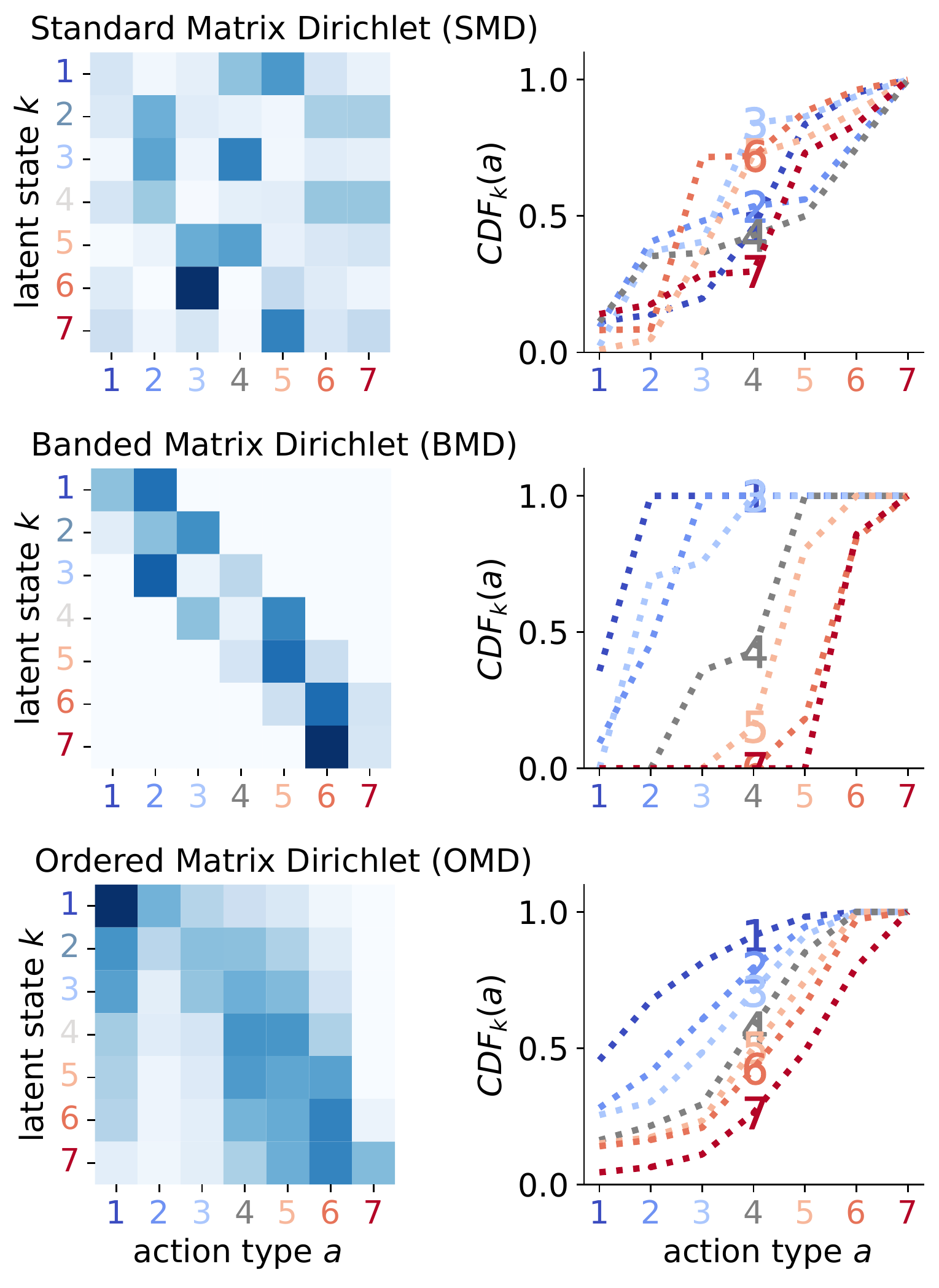}
\caption{Stochastic matrices sampled from the \SMD (\smd), \BMD (\bmd) and \OMD (\omd). Neither the \smd nor the \bmd adhere to the stochastic dominance property in~\cref{eq:fsd_cdf}, as evidenced by overlapping CDFs.\looseness=-1}
\label{fig:fsd}
\end{figure}

\paragraph{Banded Constraints}
One commonly-used constraint is that the transition matrix is \defn{banded} along its diagonal, such that $\pi_{k'k}=0$ if $|k-k'| > b$ for some bandwidth $b$ (often set to 1); see the middle plot of \cref{fig:fsd}. This constraint encodes the assumption that states only excite or transition to nearby states at subsequent time steps. Such an assumption is motivated, for example, when the desired state space represents ordered stages of escalation in international conflict~\citep{schrodt_forecasting_2006, randahl_predicting_2022}. For purposes of comparison, we introduce a prior distribution called the \defn{\BMD (\bmd)} that enforces this constraint (\cref{sec:bmd_details}).\looseness=-1

\section{THE ORDERED MATRIX DIRICHLET}
\label{sec:omd}

Many dynamical systems in the real world are naturally described by latent states with intrinsic ordering, such as in international relations, where the relationship status of two countries might escalate from \setquote{ally} to \setquote{enemy} only gradually, first passing through intermediate states like \setquote{neutral}. In addition to constraining how states transition over time, this ordering may further reflect ordering in the observed actions between countries, with countries in more conflictual latent states (e.g., \setquote{enemy}) taking more conflictual actions towards each other (e.g., \setquote{fight}), and countries in more cooperative states (e.g., \setquote{ally}) taking more cooperative actions (e.g., \setquote{provide aid}).\looseness=-1

The state space in the basic model formulation given in \cref{eq:emission} and~\cref{eq:transition} is not intrinsically ordered. Specifically, the row indices $k \in [K]$ of the emission and transition matrices are arbitrary and bear no intrinsic information. As mentioned in the previous section, constraining the transition matrix to be banded does impart some information to the index $k$. However, its interpretation is circular: $k$ is some state that transitions to other states $\{k': |k-k'| \leq b\}$, whose interpretation is similarly defined with respect to $k$. Moreover, while banding the transition matrix promotes some ordering of latent states, it does not promote one that necessarily reflects the ordering in observed actions.

To overcome these limitations, this section introduces a novel prior over the transition and emission matrices that ensures an intrinsically well-ordered state space, one that both reflects the ordering in observed actions and the ordering in latent state transitions. We first operationalize our notion of ordering in terms of stochastic dominance, and then construct a probability distribution with support over the subset of stochastic matrices that obey this notion.\looseness=-1

\subsection{Ordering by Stochastic Dominance} 

Considering first the emission matrix, each row $\boldsymbol{\phi}_k$ represents a discrete distribution over ordinal actions types. Intuitively, we might say the rows are well-ordered if probability mass shifts to the right when moving down rows, or equivalently, when the $k^{\textrm{th}}$ distribution places more weight on earlier action types than the $(k \tp 1)^{\textrm{th}}$. This intuition is formalized by the notion of \emph{stochastic dominance}~\citep{Davidson2017}. Define the cumulative distribution function (CDF) for the $k^{\textrm{th}}$ discrete distribution to be $\mathrm{CDF}_k(a) \triangleq \sum_{a'=1}^a \phi_{ka'}$. Then, the $k^\textrm{th}$ distribution is stochastically dominated by the $(k \tp 1)^\textrm{th}$ if\looseness=-1
\begin{equation}
\label{eq:fsd_cdf}
\mathrm{CDF}_k(a) \geq \mathrm{CDF}_{k \tp 1}(a) \,\,\textrm{ for all } a
\end{equation}
We refer to a stochastic matrix as \emph{well-ordered} if~\cref{eq:fsd_cdf} holds for all rows $k$. For the $K \times A$ emission matrix, this means higher rows place more mass on earlier actions types. For the $K \times K$ transition matrix, this notion encapsulates many but not all banded structures (see~\Cref{fig:fsd}), while further allowing for much more flexible \setquote{down-and-right} transition shapes (see~\Cref{fig:synth_forecast_hmm}).~\looseness=-1

\begin{figure*}[t]
     \centering
     \includegraphics[width=1.0\linewidth]{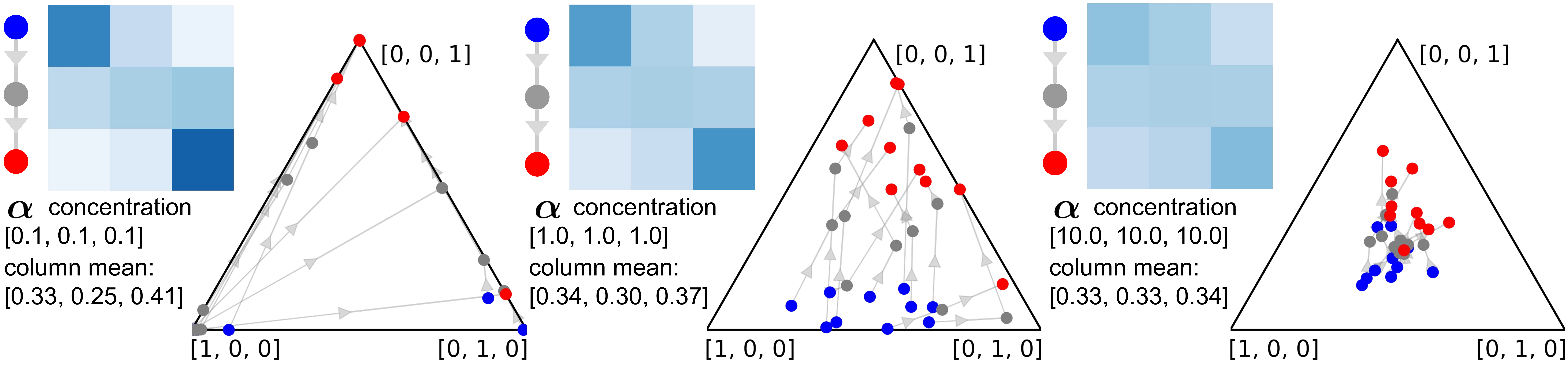} 
     \caption{\emph{Heatmaps:} Averages of 10 samples from the \OMD  with varying concentration $\valpha \in \mathbb{R}_+^{A}$. \setquote{Column mean} refers to $\overline{\phi_{a}} = \frac{1}{K}\sum_{k=1}^{K}\vphi_{ka}$ and shows that probability mass is asymmetric to the right. \emph{Simplex plots:} Each point is a discrete distribution over $A=3$ classes. The $K = 3$ points (\tikzcircle[blue, fill=blue]{2.5pt}, \tikzcircle[grey, fill=grey]{2.5pt}, \tikzcircle[red, fill=red]{2.5pt}) connected by a line represent one sample from the \omd. We observe ordered transitions from the lower left $[1,0,0]$ to the top $[0,0,1]$ corner of the simplex. In \cref{fig:smd_samples} in \cref{sec:plots}, we present unordered sample trajectories from the \smd in comparison.}
     \label{fig:prior_analysis}
\end{figure*}

\subsection{Ordered Matrix Dirichlet (\omd) Distribution}
\label{sec:omd_construction}

We now introduce a new probability distribution that has support over only the matrices described above. The \omd distribution is defined by two parameters, \emph{concentration} $\valpha \in \mathbb{R}_+^{A}$ and \emph{height} $K$. An \omd random variable $\Phi \sim \textrm{\omd}(K, \valpha)$ is a $K \times A$ matrix that is row-stochastic and well-ordered, as shown in~\Cref{prop:order}.~\looseness=-1 

We define the \omd distribution implicitly via~\Cref{alg:omd}, which generates OMD variates. This algorithm builds on the stick-breaking construction of the standard Dirichlet distribution~\citep[p.~583]{gelman_bayesian_2013}. A Dirichlet random variable $\phi \sim \textrm{Dir}(\boldsymbol{\alpha})$ can be generated iteratively, one entry at a time, via Beta auxiliary variables. First, draw $\phi_{1} \sim \textrm{Beta}(\alpha_1, \sum_{a>1} \alpha_a)$. Then for $a = 2,\dots,A \tm 1$ draw $\beta_{a} \sim \textrm{Beta}(\alpha_a, \sum_{a'>a} \alpha_{a'})$ and set $\phi_a \leftarrow \beta_a \, (1-\sum_{a'< a} \phi_{a'})$. Finally, set $\phi_A \leftarrow 1 - \sum_{a'<A} \phi_{a'}$. Intuitively, this construction iteratively \setquote{breaks} off some amount of remaining probability mass (the \setquote{stick}), where the Beta variables determine the size of the breaks.

\Cref{alg:omd} iteratively constructs $K$ discrete distributions over $A$ categories using the same basic idea. For each category $a$ in succession, it samples $K$ Beta variables (lines 3 and 8) to determine the size of the \setquote{breaks} in the $K$ \setquote{sticks}. However, it further sorts the Beta variables (lines 5 and 10), so that the largest \setquote{break} of the remaining \setquote{stick} is always taken by the first \setquote{stick} ($k=1$), the second largest is always taken by the second \setquote{stick} ($k=2$), and so on. In so doing, it generates a well-ordered stochastic matrix, as stated below.\looseness=-1

\begin{proposition}[\omd random variables are well-ordered]
The \omd has support over only row-stochastic matrices that obey the ordering property given beneath~\Cref{eq:fsd_cdf}, such that for any two rows $k  < k'$ and any $a$
\begin{equation}
    \sum_{a'\leq a} \phi_{ka'} \geq \sum_{a'\leq a} \phi_{k'a'}
\end{equation}
\textbf{Proof:} See~\Cref{sec:proof}.
\label{prop:order}
\end{proposition}

\paragraph{Lack of Analytic Form} 
We define the \omd implicitly by construction and do not (yet) know any analytic form for its probability density function (PDF), which involves integrating over products of Beta order statistics. We leave further investigation into the \omd's PDF, moments, and other analytic properties for the future. As we show in the next section though, its lack of analytic form does not hamper posterior inference with modern probabilistic programming.~\looseness=-1

\paragraph{What is \setquote{Dirichlet} about the \omd?} The \omd's name reflects its definition as a minimal modification to the stick-breaking construction of the \SMD---if we remove the blue lines (5 and 10), then~\Cref{alg:omd} corresponds exactly to the \smd, which simply generates $K$ independent Dirichlet variates (with no ordering). The name reflects this alone---it is not the case (to our knowledge) that the $K$ discrete distributions, which are dependent under the \omd via the sort operation, are marginally or conditionally Dirichlet distributed.

\begin{algorithm}[t]
\caption{\OMD}
\label{alg:omd}
\begin{algorithmic}[1]
\State \textbf{Input:} height $K$, concentration $\valpha \in \mathbb{R}_+^{A}$
\For{$k = 1,\,\dots\,, K$}
\State $\tilde{\vphi}_{k1} \sim \textrm{Beta}\left(\alpha_1, \sum_{a=2}^A \alpha_a\right)$
\EndFor
\color{blue}
\State $(\phi_{11},\dots,\phi_{K1}) \leftarrow \textsc{Sort}\Big((\tilde{\phi}_{11},\dots,\tilde{\phi}_{K1})\Big)$
\color{black}
\For{$a = 2,\,\dots\,,\, A-1$}
\For{$k = 1,\,\dots\,, K$}
\State $\tilde{\beta}_{ka} \sim \textrm{Beta}\left(\alpha_a, \sum_{a'=a+1}^A \alpha_{a'}\right)$
\EndFor
\color{blue}
\State $(\beta_{1a},\dots,\beta_{Ka}) \leftarrow \textsc{Sort}\Big((\tilde{\beta}_{1a},\dots,\tilde{\beta}_{Ka})\Big)$ \color{black}
\For{$k = 1,\,\dots\,, K$}
\State $\vphi_{ka} \leftarrow \left(1 - \sum_{a'=1}^{a-1} \phi_{ka'}\right) \beta_{ka}$
\EndFor
\EndFor
\For{$k = 1,\,\dots\,, K$}
\State $\vphi_{kA} \leftarrow 1 - \sum_{a'=1}^{A-1} \phi_{ka'}$
\EndFor
\State \textbf{Output:} \omd variate $\Phi \in \mathbb{R}_+^{K \times A}$
\end{algorithmic}
\end{algorithm}

\paragraph{Concentration Parameter}

The \omd is parameterized by its concentration $\valpha \in \mathbb{R}_+^{A}$. \cref{fig:prior_analysis} visualizes OMD samples $\Phi$ for different settings of $\valpha$. For symmetric $\valpha=(\alpha_0,\dots,\alpha_0)$, one might expect samples $\Phi$ to distribute probability mass across the matrix evenly. However, we observe otherwise, that mass often skews to the right, particularly for larger $\alpha_0$; we speculate this relates to the sorting operation. Although unappealing, this does not mean the \omd is inherently asymmetric, but rather that non-trivial settings of $\valpha$ may be required to promote symmetry in the prior. Despite this, in practice, we find that samples from the posterior are often symmetric.~\looseness=-1

\paragraph{Label Switching}
The problem of \setquote{label switching} \citep[p.~841]{stephens_dealing_2000, murphy_machine_2012} arises in (ad)mixture models when the indices $k$ of latent states are arbitrary, such that permuting them gives the same joint probability under the model. This issue can hamper interpretation and prevents one from averaging parameters across posterior samples without first aligning the states (e.g., using the Hungarian matching algorithm~\citep{kuhn_hungarian_1955}). The models we have discussed, which place an \omd prior over the emission matrix, have intrinsically ordered states that are not prone to label switching (see \cref{fig:dir_label_switching} in \cref{sec:plots}). Although we do not view this as the main benefit of the \omd, it is a welcome side effect that facilitates easier interpretation and permits direct averaging of posterior parameters without post-hoc, potentially error-prone alignment methods.

\section{MCMC INFERENCE WITH PYRO}
\label{sec:inference}
The \omd integrates nicely with modern probabilistic programming frameworks like \defn{\href{https://docs.pyro.ai/en/stable/}{Pyro}}~\citep{bingham_pyro_2018, phan_composable_2019}.\footnote{We open-source our code with tutorials and examples at \href{\link}{\link}} Although we do not have an analytic form for its PDF, we are able to build and perform efficient gradient-based MCMC on a range of OMD-based models by implementing~\cref{alg:omd}. We regard the \omd as a modeling motif that blends white- and black-box approaches in a way that was only recently made feasible by advances in scientific computing.~\looseness=-1

We use Pyro's implementation of the \href{https://docs.pyro.ai/en/stable/_modules/pyro/infer/mcmc/nuts.html}{No-U-Turn Sampler} \citep[NUTS,][]{hoffman_no-u-turn_2014}, a variant of Hamiltonian Monte Carlo \citep[HMC,][]{duane_hybrid_1987}, to perform approximate posterior inference in \omd-based models. As with any MCMC method, this returns a set of $S$ posterior samples of model parameters $\{\Pi^{(s)},\Phi^{(s)},\dots\}_{s=1}^S$ which collectively approximate the posterior distribution. In practice, we take $S = $\num{1000} samples after \num{200} burn-in samples.

NUTS relies on first-order gradient information of the model's unnormalized log joint density. Our implementation in Pyro takes gradients of the \omd density implicitly via backpropagation through the stick-breaking construction. Although the \texttt{sort} operation is not fully differentiable, it is piece-wise linear and sub-differentiable \citep{boyd_convex_2004, blondel_fast_2020, tim_vieira_distribution_2021}. We can view \texttt{sort} as a combination of two operations: first the non-differentiable \texttt{argsort} obtains permutation indices, then the differentiable \texttt{gather} applies the permutation. In the backward pass, the permutation of indices is simply reversed to match their original positions, obviating the need to differentiate through \texttt{argsort}. For this reason, the sorting of Beta variates in the construction of the \omd does not hinder gradient-based MCMC methods for inference.~\looseness=-1

\begin{figure*}[t]
     \centering
     \includegraphics[width=1.0\linewidth]{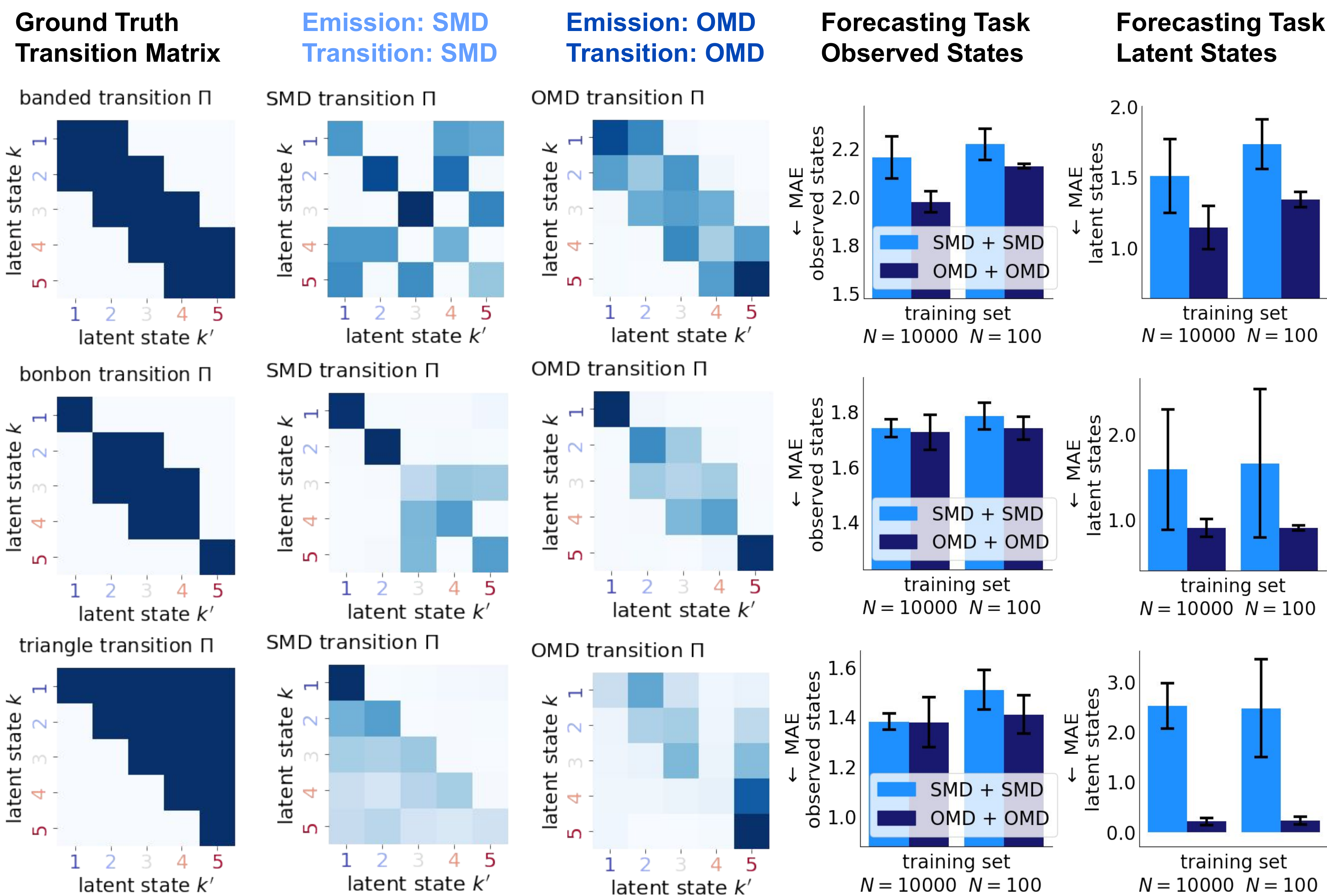}
     \caption{\smd versus \omd at forecasting. \emph{Column 1}: Stylized ground-truth transition matrices. \emph{Columns 2-3}: \omd recovers the transition matrices while the \smd suffers from label switching. \emph{Columns 4:} \omd is better at forecasting than \smd but worse at imputation. \emph{Columns 5:} \omd recovers the latent states while the \smd suffers from label switching.}
     \label{fig:synth_forecast_hmm}
 \end{figure*}

\section{SYNTHETIC DATA EXPERIMENTS}
\label{sec:synthetic_data}
We conduct experiments with synthetic data to better understand and evaluate the behavior of \ssms with \omd priors. In particular, we generate datasets using hidden Markov models (HMMs) with roughly diagonal emission matrices and a range of stylized transition matrices---i.e., \setquote{banded}, \setquote{bonbon} and \setquote{triangle}, all displayed in the left column of \cref{fig:synth_forecast_hmm}. The \setquote{bonbon}, for instance, represents a realistic scenario for political event data, where \setquote{neutral} states fluctuate but \setquote{ally} and \setquote{enemy} states are nearly absorbing. To each of these datasets, we fit an HMM with \omd priors and compare its performance to a baseline HMM with \smd priors.~\looseness=-1

We generate multiple datasets for each parameter setting using \num{10} random seeds, where each dataset comprises $N=$ \num{10000} sequences of length $T=10$, and where a single observation takes one of $A=10$ ordinal values. We further consider two settings: one with all $N =$ \num{10000} sequences and a \defn{few-shot} setting where models are fit to only $N = 100$. We also generate random train-test splits to evaluate two different forms of prediction: 

\begin{enumerate}[wide = 0pt, label=({{\arabic*}})]
\item \defn{Imputation}: We mask a random \num{30}\% of all observations which models impute during inference.
\item \defn{Forecasting}: We designate the first \num{70}\% of time steps for training and the latter \num{30}\% for testing. Models are fit to the training set, then used to forecast the test observations.
\end{enumerate}

\paragraph{Qualitative Results.} 
We first compare how well the two models recover known ground-truth latent structure. As expected, the \omd model reliably recovers the shape of the true transition matrix while the \smd model does not, often exhibiting label switching; see the first 3 columns of~\cref{fig:synth_forecast_hmm} for examples. As a simple quantitative measure of this, we can also calculate the mean absolute error (MAE) between the true latent states and the inferred ones. The last column of~\cref{fig:synth_forecast_hmm} reports the error on forecasting future latent states where the \omd model is substantially better; this is unsurprising and simply confirms that the \omd's states are well-ordered while the \smd's are label-switched.

\paragraph{Predictive Results.} The $4^{\textrm{th}}$ column of~\cref{fig:synth_forecast_hmm} reports MAE on forecasting future observations. The \omd model performs at least as well as the \smd model in all settings, and sometimes substantially better, as when the true transition matrix is banded ($1^{\textrm{st}}$ row). By contrast, the imputation results in~\cref{fig:synth_impute_hmm} in \cref{sec:plots} show the \omd model performing substantially worse than the \smd model in most settings. We speculate that the strong inductive bias imparted by the \omd prior helpfully regularizes the model's forecasts while overly restricting its imputation ability. Intriguingly, the one setting where the \omd model has superior imputation performance is when the true transition matrix is banded, which accords with the forecasting results.

\begin{figure}[t]
     \centering
     \includegraphics[width=\linewidth]{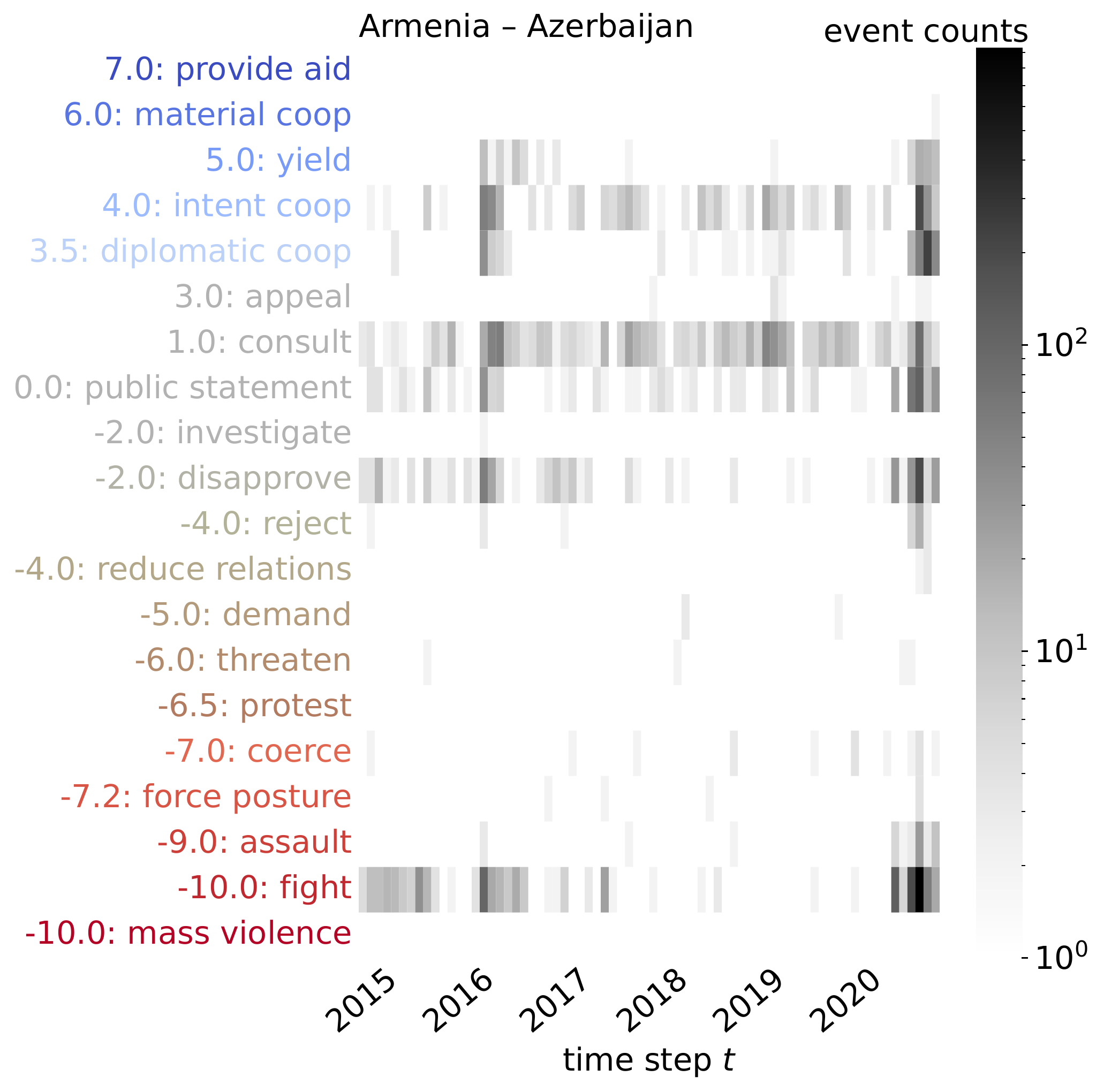} 
     \caption{ICEWS event data showing interactions between \textsc{Armenia} and \textsc{Azerbaijan} over monthly time steps.}
     \label{fig:count_data_month}
 \end{figure}

\begin{figure}[t]
\centering
\includegraphics[width=1.0\linewidth]{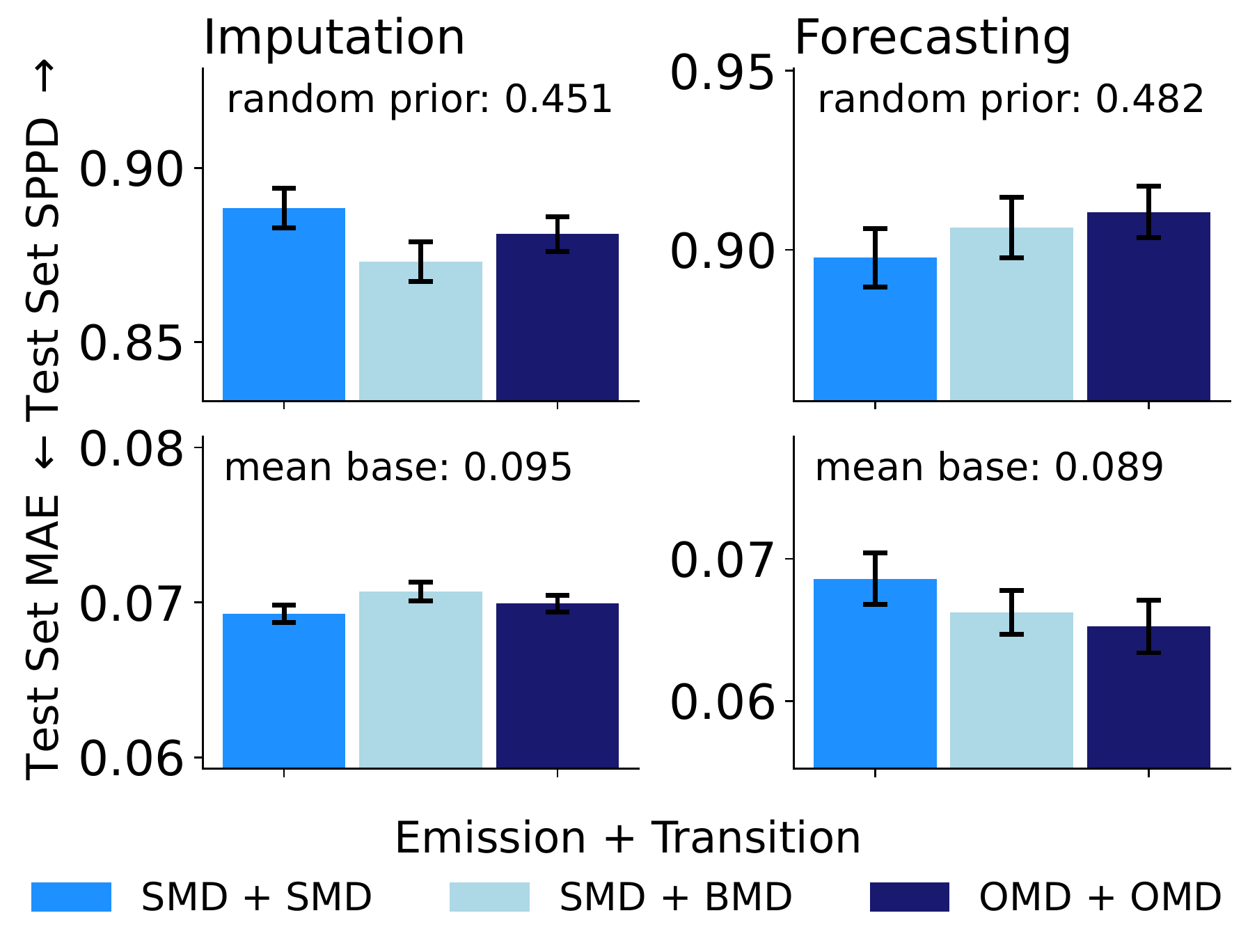}
\caption{Imputation and forecasting evaluation on held-out ICEWS data, over \num{10} runs with random seed. We fit the \dpt model with different parametrizations (\smd, \bmd, \omd) of the emission $\Phi$ and transition $\Pi$ matrix. We find that \omd does not significantly reduce predictive results suggesting that the imposed constraints fit the given data.}
\label{fig:real_predictive} 
\end{figure}

\begin{figure*}[t]
 \centering
 \includegraphics[width=1.0\linewidth]{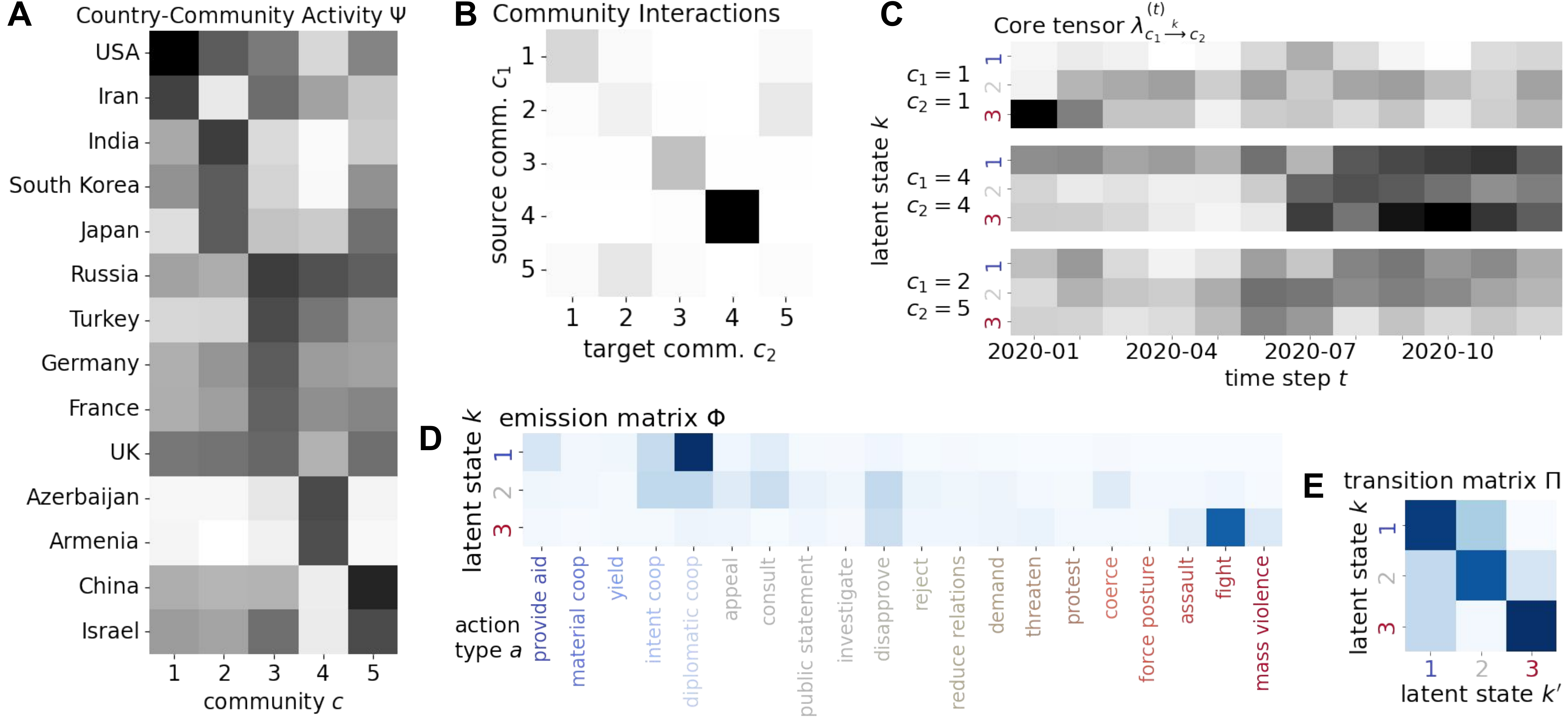}
 \caption{Posterior mean of parameters of \DPTM fitted to ICEWS subset of 2020. (A) The latent matrix $\Psi$ indicates country-community activity. \textsc{Armenia} and \textsc{Azerbaijan} are standing out as they are involved mostly in one community (B) Interactions between latent communities. We find that communities $c=1$, $c=3$ and $c=4$ predominantly interact between themselves. (C) Selected community interactions over time: $c_{1} = 4 \rightarrow c_{2} = 4$ are in conflict, while $c_{1} = 2 \rightarrow c_{2} = 5$ are mostly neutral. (D) Latent emission matrix $\Phi$ representing global state-to-action probabilities. Thanks to the ordering, we know that state $k=3$ represents conflictual relationships. (E) Latent transition matrix $\Pi$ representing smooth state-to-state transition probabilities.
 }
 \label{fig:real_results_20}
\end{figure*}

\section{CASE STUDY: POLITICAL EVENTS}
\label{sec:application}

In this section, we give a case study on building an \omd-based \ssm to analyze international relations event data.\looseness=-1 

\subsection{ICEWS Political Events Data}
We consider political event data from the \href{https://dataverse.harvard.edu/dataverse/icews}{Integrated Crisis Early Warning System (ICEWS)} dataset \citep{boschee_icews_2015}. ICEWS event data comprise millions of micro-records of the form ``country $i$ took action $a$ to country $j$ at time $t$'' that are machine-extracted from digital news archives. The country actors $i$ and $j$ and action types $a$ are coded to follow the \href{https://parusanalytics.com/eventdata/cameo.dir/CAMEO.09b6.pdf}{Conflict and Mediation Event Observations (CAMEO)} ontology \citep{ schrodt_cameo_2012}.

\paragraph{Ordered Actions} CAMEO specifies \num{20} high-level action types, depicted in \cref{fig:count_data_month}, that are naturally ordered on a conflictual-to-cooperative axis---specifically, they are each assigned a value on the expert-elicited \href{https://parusanalytics.com/eventdata/cameo.dir/CAMEO.SCALE.txt}{Goldstein scale} \citep{goldstein_conflict-cooperation_1992}, where  the most cooperative action, \setquote{provide aid}, has a value of $+7.0$, and the most conflictual action, \setquote{use unconventional mass violence}, has a value of $-10.0$.\looseness=-1

\paragraph{4-mode Count Tensor}

Following \citet{schein_bayesian_2016}, we represent the data as a count tensor $\tY \in \mathbb{N}_{0}^{\bV \times \bV \times A \times T}$, where an element $\yijat$ is the number of times country $i$ took action $a$ to country $j$ during time step $t$. We consider $V = 249$ countries, $A = 20$ action types (ordered by Goldstein values), and $T = 72$ months. The $A \times T$ slice of this tensor corresponding to all interactions between $i \equiv \textsc{Armenia}$ and $j \equiv \textsc{Azerbaijan}$ is visualized in \cref{fig:count_data_month}.\looseness=-1

\subsection{\DPTM}
\label{sec:dptm}

Our model assumes each count $\yijat$ is Poisson distributed:
\begin{align}
    \yijat &\sim \textrm{Pois} \left( \delta_{a} \, \delta^{(t)} \sum_{k=1}^K \lambdaijkt \underbrace{\phi_{ka}}_\text{\tiny emission} \right)
    \label{eq:emission_dptm}
\end{align}
where $\phi_{ka}$ is an entry in the state-to-action emission matrix, the parameters $\delta_{a}$ and $\delta^{(t)}$ are action- and time-scaling coefficients, and $\lambdaijkt$ represents how well the $k^{\textrm{th}}$ state describes the relationship $(i \rightarrow j)$ at time $t$. \Cref{eq:emission_dptm} conforms to the basic form given in~\Cref{eq:emission}, where here the measurements and states are specifically tensor-valued.

Our model further assumes that $\lambdaijkt$ decomposes so that 
\begin{align}
   \sum_{k=1}^K \lambdaijkt \phi_{ka} \equiv \sum_{c_1=1}^C \psi_{c_1i} \sum_{c_2=1}^C \psi_{c_2j} \sum_{k=1}^K  \lambda^{(t)}_{c_1 \xrightarrow{k} c_2}\phi_{ka} 
   \label{eq:tucker}
\end{align}
where $\psi_{c_1i}$ and $\psi_{c_2j}$ represent the rate at which countries $i$ and $j$ participate in \emph{latent communities} $c_1$ and $c_2$, respectively, and $\lambda^{(t)}_{c_{1} \xrightarrow{k} c_{2}}$ then represents how well the $k^{\textrm{th}}$ state describes the inter-community relationship $(c_1 \rightarrow c_2)$ at time $t$. The multilinear form in~\Cref{eq:tucker} corresponds to a Tucker decomposition~\citet{tucker_extension_1964}, where the $\lambda^{(t)}_{c_{1} \xrightarrow{k} c_{2}}$ values collectively form the \defn{core tensor} $\tLambda^{(t)} \in \mathbb{R}_+^{C \times C \times K}$ at time $t$. In this setting, the core tensor can also be interpreted as a tensor-valued \emph{state (of the whole system)}.

We then model the evolution of the core tensor over time as
\begin{align}
    \lambda^{(t)}_{c_1 \xrightarrow{k} c_2} &\sim \textrm{Gam}\left(\tau_0 \,\sum_{k'=1}^K \lambda^{(t-1)}_{c_1 \xrightarrow{k'} c_2} \underbrace{\pi_{k'k}}_\text{transition},\, \tau_0\right)
    \label{eq:transition_dptm}
\end{align}
which follows the form of Poisson--Gamma Dynamical Systems~\citep{schein_poisson-gamma_2016} while conforming to~\Cref{eq:transition}. 

We place non-informative gamma priors over the parameters $\delta_a,\, \psi_{c_1i},\,\psi_{c_2j} \stackrel{\textrm{iid}}{\sim} \textrm{Gam}(\alpha_0, \alpha_0)$, a dynamic prior over $\delta^{(t)} \sim \textrm{Gam}(\tau_0\delta^{(t \tm 1)},\,\tau_0)$, and set $\tau_0=\alpha_0=1$.

Finally, with all of the aforementioned structure the same, we then consider three different settings for the priors over the transition $\Pi$ and emission $\Phi$ matrices: (1) \defn{OMD+\omd}, where both are drawn from the \omd, (2) \defn{SMD+\smd}, where both are drawn from the \smd, and (3) \defn{SMD+\bmd}, where $\Phi$ is drawn from the \smd and $\Pi$ is drawn from the \BMD (\bmd), as defined in~\Cref{sec:bmd_details}. 

\subsection{Experiments and Results}
\label{sec:real_data}

To further understand and evaluate the \omd we fit the three above-mentioned versions of the \DPT (\dpt) model to ICEWS data and compare their qualitative and predictive performance. We use the same hyperparameters for all models with $C=5$ and $K=3$. 

\paragraph{Predictive Evaluation}
Following the design in~\cref{sec:synthetic_data}, we create \num{10} train-test splits that randomly mask observations for imputation and withhold later time steps for forecasting. In addition to MAE, we evaluate performance using \emph{scaled pointwise predictive density (\sppd)}, a measure between \num{0} and \num{1} where higher is better, which we define in~\Cref{sec:sppd_details}.~\looseness=-1 

\Cref{fig:real_predictive} reports the imputation and forecasting  results for each of the three models. As in the synthetic experiments, we see that the \omd model is better than the \smd at forecasting but worse at imputation. Similarly, the \bmd model is also better than the \smd at forecasting but worse at imputation. This strengthens our belief that the \omd's inductive bias regularizes its forecasts while overly restricting its imputation ability, since the \bmd exhibits the same pattern, and their two inductive biases are similar. That being said, the \omd is much more flexible than the \bmd, which may explain why it outperforms the \bmd in both forecasting and imputation.\looseness=-1 

\paragraph{Qualitative Exploration}
To qualitatively inspect its inferred latent structure, we fit the \omd model to the fully-observed dataset. \Cref{fig:real_results_20} visualizes the \emph{posterior mean} of inferred model parameters for the time period of 2020. Since the model is not prone to label switching, we can inspect the posterior mean, as opposed to inspecting single (often arbitrary) sample. \cref{fig:real_results_20}A visualizes the country-community matrix $\Psi$. We observe that \textsc{Armenia} and \textsc{Azerbaijan} are predominantly involved in community $c = 4$. By visualizing a slice of the core tensor in~\cref{fig:real_results_20}B, we see that this community mostly interacts with itself. By visualizing in~\cref{fig:real_results_20}C the slice $\lambda^{(t)}_{4 \xrightarrow{k} 4}$, we see which states $k$ best describe community $c=4$'s self-interactions over time. In mid 2020, the most active state is $k=3$. We immediately know it represents a conflictual relationship since its index $k$ is high. This is confirmed by the emission matrix in~\cref{fig:real_results_20}D where we see that state $k=3$ places most of its mass on \setquote{fight}. Finally, we visualize the transition matrix in~\cref{fig:real_results_20}E and find that, unfortunately, transitioning out of state $k=3$ seems unlikely.\looseness=-1

We also visualize inferred latent structure from a \dpt model with $K=6$ and $C=20$ fitted to ICEWS data from a longer time range 2015--2020 and present the results in~\cref{fig:real_results_15_20}.

\section{DISCUSSION}
\label{sec:discussion}

\paragraph{International Relations} As alluded to throughout, this work was largely motivated by datasets, modeling approaches, and core concepts in the field of international relations (IR). The notion of \emph{escalation}---that countries only gradually transition to conflict through an orderly sequence of intermediate states---is fundamental to how scholars organize and understand political events~\citep{davis_concepts_1984}. Theoretical accounts for why countries fight attempt to characterize a sequence of intermediate states that rational actors would transition through on their way to war~\citep{snyder1984security,fearon1995rationalist,jervis2017perception}. A similar perspective underlies empirical approaches. The earliest attempts to digitize international affairs into \setquote{event data} were explicitly couched in the framework of escalation---the very first sentence of~\citet{azar1980conflict} reads: ``As students of politics and political science, we should and we do care about the events which lead to war...''\looseness=-1

The principal challenge in the data-intensive study of international relations is the inherent sparsity and missingness of event data, which provide only a scattered glimpse at the underlying structures we seek to reason about. This paper follows an empirical tradition of encoding strong inductive biases into statistical models of event data which encourage their inferred structure to accord with theoretical notions, like \setquote{escalation}~\citep{schrodt_forecasting_2006, anders_territorial_2020, randahl_predicting_2022}. While much of the previous work focuses on constraining (specifically, banding) the transition structure between ``states'' to encourage orderly dynamics, the key idea in this paper is to draw further on the ordinal nature of observed action types. There is a steadily-growing literature on models for dyadic event data that has made exciting advances while still mostly treating action types as unordered~\citep{oconnor_learning_2013, schein_bayesian_2015, minhas_new_2016}. In parallel, there has been recent work on inferring latent intensity scales~\citet{terechshenko_hot_2020,stoehr_ordinal_2022} that imbue actions with a richer or more data-driven sense of ordering. We are eager for these threads to continue to cross, as they have in this work.

\paragraph{Other Models and Other Domains}

The \OMD as a modeling motif is applicable beyond international relations and \ssms. We include in~\Cref{sec:plots} a brief exploration of other \omd-based models we have built, with illustrative results on other datasets. Building these models in Pyro is easy, often only requiring a few lines of code, which facilitates this exploration. \Cref{fig:other_models} summarizes four different models, all of which (and more) are available in the code we have open-sourced; we describe them here too.\looseness=-1

\begin{enumerate}[wide = 0pt, label=({{\arabic*}})]
\item We build an ordered form of Poisson--Gamma Dynamical Systems (PGDS) \citep{schein_poisson-gamma_2016} placing an \omd prior over the transition and emission matrices. PGDS was originally introduced to model ICEWS data, but treats actions as unordered.\looseness=-1

\item We use an HMM with \omd-distributed emission and transition matrices to model the observed global change of temperature. In this model, noisy temperature changes are related to ordered latent states indicative of \setquote{warming} and \setquote{cooling} periods that transition gradually.

\item Even simpler, we experiment with a Markov chain model consisting of a single state-to-state transition matrix to model sleep cycles. Sleep cycles typically transition step-by-step from wake (W) to rapid eye movement (REM) stages \citep{pan_transition-constrained_2012}. 

\item We modify Latent Dirichlet Allocation (LDA)~\citep{blei_latent_2003} to place an \omd prior over the topic-word matrix. While word types are canonically viewed as unordered, we imbue them with ordering by sorting them on \setquote{semantic axes} \citep{an_semaxis_2018}, for instance from negative to positive words. The model then infers ordered topics that reflect this semantic axis, similar to the model of~\citet{stoehr_sentiment_2023}.
\end{enumerate}

We can imagine many more applications that motivate well-ordered state-space models, like modeling product life cycles~\citep{arvidsson_use_2019} or customer-company relationships~\citep{netzer_hidden_2008}. Beyond \ssms, admixture models, like LDA, are fundamentally based on stochastic matrices and used in population genetics~\citep{pritchard2000inference}, stochastic block models~\citep{airoldi2008mixed}, recommender systems~\citep{gopalan2015scalable}, among many other areas.\looseness=-1

\section{CONCLUSION}
\label{sec:conclusion}
This paper introduced the \OMD (\omd) distribution as a prior distribution over well-ordered stochastic matrices in state-space models (\ssms). Models built on the \omd have intrinsically ordered states that reflect ordering in the observed data. These models are more readily interpretable and usable, as they are not prone to label switching, while still being competitive on predictive tasks. The \omd integrates nicely with modern probabilistic programming frameworks, making it easy to build and fit \omd-based models. While this paper's motivation is rooted in the concepts and data of international relations, the motifs presented here have broad applicability to domains with ordinal data and models based on stochastic matrices.

\subsubsection*{Acknowledgments}
We would like to thank Kevin Du and the anonymous reviewers for valuable feedback on the manuscript and Tim Vieira for his input on \href{https://timvieira.github.io/blog/post/2021/03/18/on-the-distribution-functions-of-order-statistics/}{order statistics and sorting}. Niklas Stoehr is supported by the Swiss Data Science Center (SDSC).

\bibliography{refs/nik,refs/aaron}

\clearpage
\appendix

\thispagestyle{empty}

\onecolumn

 \section{IMPACT STATEMENT}
\label{sec:impact}

We emphasize that our models are intended for research purposes and empirical insights. They should not be blindly deployed for automated decision-making processes. The used ICEWS data may contain biases that are potentially reinforced by our modeling assumptions. The experiments with real-world event data in \cref{sec:real_data} were conducted on an NVIDIA TITAN RTX GPU. The experiments with synthetically generated data in \cref{sec:synthetic_data} can be run on a local M1 CPU with \num{64} GB of RAM in less than \num{10} minutes. Limiting factors are the selected hyperparameter sizes for the latent states $K$ and communities $C$, as well as the number of time series $N$ and their length $T$. We discuss further model limitations in \cref{sec:conclusion} and \cref{sec:omd}.

\section{SUPPLEMENTARY TECHNICAL DETAILS}

\subsection{Proof of~\Cref{prop:order}}
\label{sec:proof}
\begin{proposition}[\omd random variables are well-ordered]
The \omd has support over only row-stochastic matrices that obey the ordering property given beneath~\Cref{eq:fsd_cdf}, such that for any two rows $k  < k'$ and any $a$
\begin{equation}
    \sum_{a'\leq a} \phi_{ka'} \geq \sum_{a'\leq a} \phi_{k'a'}
\end{equation}
\textbf{Proof:} For $a=1$, $\phi_{k1} > \phi_{k'1}$ is true by construction (line 5 of~\cref{alg:omd}). For $a=2$, by the definition in line 12, $\phi_{k2} = (1-\phi_{k1}) \, \beta_{k2}$, and therefore the CDF at $a=2$ equals $\phi_{k1} -\phi_{k1}\beta_{k2} + \beta_{k2}$. It suffices to show that $\phi_{k1} -\phi_{k1}\beta_{k2} + \beta_{k2} > \phi_{k'1} -\phi_{k'1}\beta_{k'2} + \beta_{k'2}$, since the remaining $a>2$ then follow by induction. Re-arranging terms, $(\phi_{k1} \tm \phi_{k'1}) + (\beta_{k2} \tm \beta_{k'2}) > (\phi_{k1}\beta_{k2} - \phi_{k'1}\beta_{k'2})$, which follows since we know by construction that $\beta_{k2} > \beta_{k'2}$ (line 10), and all terms $\phi_{k1},\phi_{k'1},\beta_{k2},\beta_{k'2}$ are between 0 and 1.
\end{proposition}

\subsection{Details of the \BMD (\bmd)}
\label{sec:bmd_details}

In this section, we elaborate on the \BMD (\bmd). For simplicity, we consider a square matrix $\Pi \in [0,1]^{K \times K}$, but the \bmd can be non-square as well. We assume that the $k^{\textrm{th}}$ state can only be excited by its directly neighboring states, $(k-1)^{\textrm{th}}$ and $(k+1)^{\textrm{th}}$, as well as by itself \citep{schrodt_forecasting_2006, randahl_predicting_2022}. This results in a matrix whose non-zero elements are banded along the diagonal following:
\begin{align}
\pi_{kk'} = 
\begin{cases}
\pi_k^{\mathsmaller{(\nearrow)}} &\textrm{ if } k' = k+1 \,\,\textrm{ (escalating)}\\
\pi_k^{\mathsmaller{(\searrow)}} &\textrm{ if } k' = k-1 \,\,\textrm{ (descalating)}\\
\pi_k^{\mathsmaller{(\circ)}} &\textrm{ if } k' = k \,\,\textrm{ (steady)}\\
0 &\textrm{ otherwise }
\end{cases}
\end{align}
Finally, we place a Dirichlet prior over the three non-zero elements in each $k^{\textrm{th}}$ row
\begin{align}
    (\pi_k^{\mathsmaller{(\nearrow)}}, \pi_k^{\mathsmaller{(\searrow)}}, \pi_k^{\mathsmaller{(\circ)}}) &\sim \textrm{Dir}(\alpha_0^{\mathsmaller{(\nearrow)}}, \alpha_0^{\mathsmaller{(\searrow)}}, \alpha_0^{\mathsmaller{(\circ)}})
\end{align}
Moreover, we can consider a wider bandwidth $b \geq 1$ so that components $k' \in \{k-b,\dots, k+b\}$ all excite $k$. An example of the full vector might then look like
\begin{align}
    \vpi_k = (0, \dots, 0,\pi_k^{\mathsmaller{(\searrow)}}, \pi_k^{\mathsmaller{(\circ)}}, 
    \pi_k^{\mathsmaller{(\nearrow)}}, 0, \dots, 0)
\end{align}

\subsection{Details on the Scaled Pointwise Predictive Density}
\label{sec:sppd_details}
\emph{Scaled pointwise predictive density (\sppd)} is defined as 
\begin{equation}
\textrm{\sppd} = \exp\Big(\tfrac{1}{|\mathcal{I}|} \sum_{\mathbf{i} \in \mathcal{I}} \log\Big[\tfrac{1}{S}\sum_{s=1}^S \textrm{Pois}\big( y_{\mathbf{i}};\, \mu_{\mathbf{i}}^{(s)}\big)\Big]\Big)
\end{equation}
where $\mathbf{i}$ is the multi-index of an entry $y_{\mathbf{i}}$ in the tensor---e.g., $\mathbf{i}=(i,j,a,t)$---and $\mathcal{I}$ is the set multi-indices corresponding to all entries in the test set. The term $\mu_{\mathbf{i}}^{(s)}$ is the Poisson rate in~\Cref{eq:emission_dptm} as given by the $s^{\textrm{th}}$ posterior sample of model parameters.
\sppd is the same as LPPD~\citep{gelman_understanding_2014}, but scaled by $\tfrac{1}{|\mathcal{I}|}$ and exponentiated so it is always between 0 and 1, where higher is better.

\subsection{Relevant Links}

\noindent \\Code accompanying this paper\\ \url{https://github.com/niklasstoehr/ordered-matrix-dirichlet}\\
Integrated Crisis Early Warning System (ICEWS)\\ \url{https://dataverse.harvard.edu/dataverse/icews}\\
Goldstein Scale\\
\url{https://parusanalytics.com/eventdata/cameo.dir/CAMEO.SCALE.txt}\\
Conflict and Mediation Event Observations (CAMEO)\\
\url{https://parusanalytics.com/eventdata/cameo.dir/CAMEO.09b6.pdf}

\section{SUPPLEMENTARY PLOTS}
\label{sec:plots}

\begin{figure*}[h]
     \centering
     \includegraphics[width=1.0\linewidth]{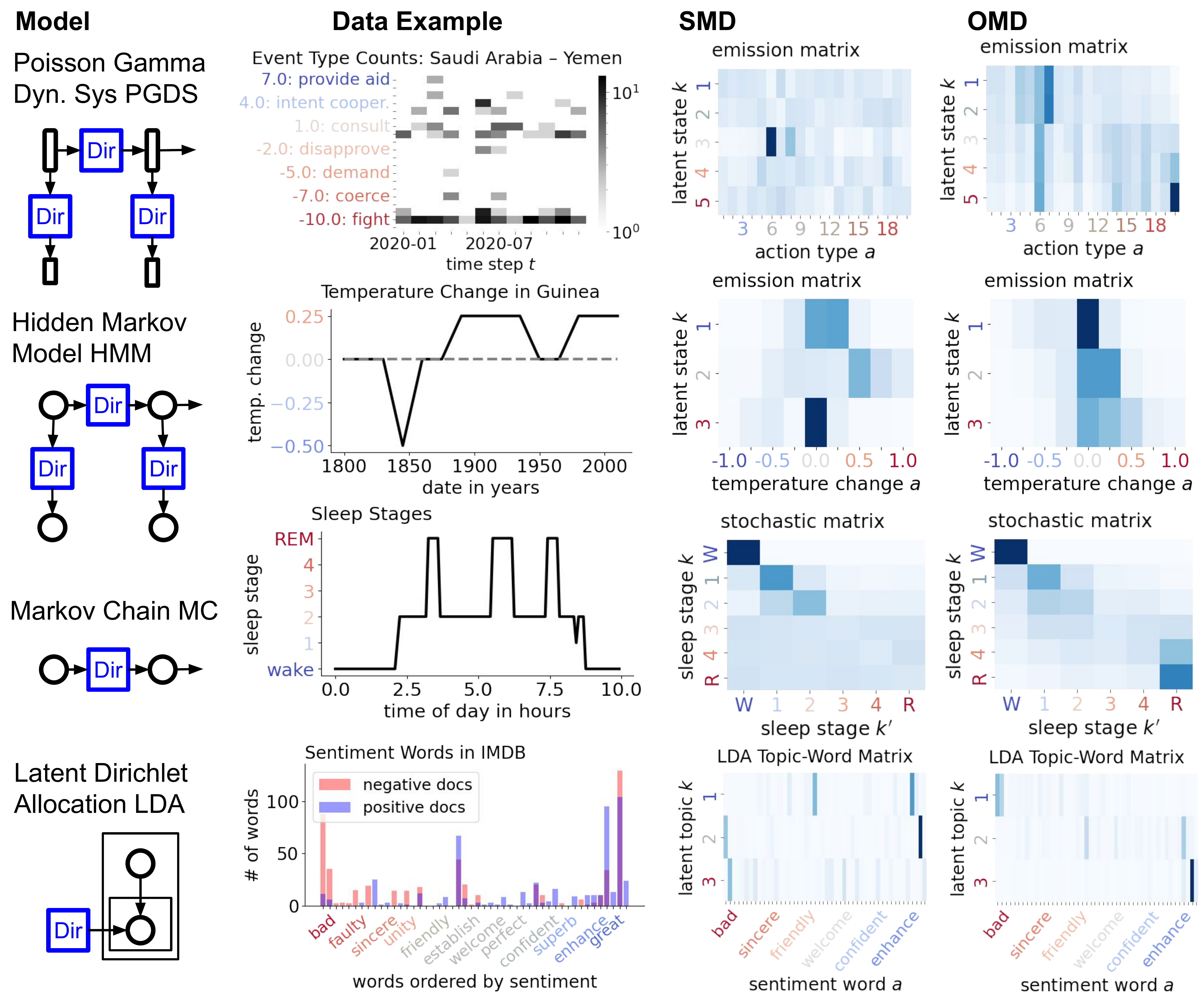} 
     \caption{Different models with Dirichlet-sampled latent matrices fitted on data exhibiting ordinal dynamics. The Latent Dirichlet Allocation (LDA) has no temporal dimension, but similarly comprises a stochastic matrix describing word distributions per latent topic. If we order the observed vocabulary of words by the words' sentiment score, the \OMD (\omd) can recover topics representative of sentiment levels. In all settings, we find that the \omd yields more easily interpretable stochastic matrices than the \SMD (\smd).}
     \label{fig:other_models}
 \end{figure*}

\begin{figure*}[h]
 \centering
 \includegraphics[width=1.0\linewidth]{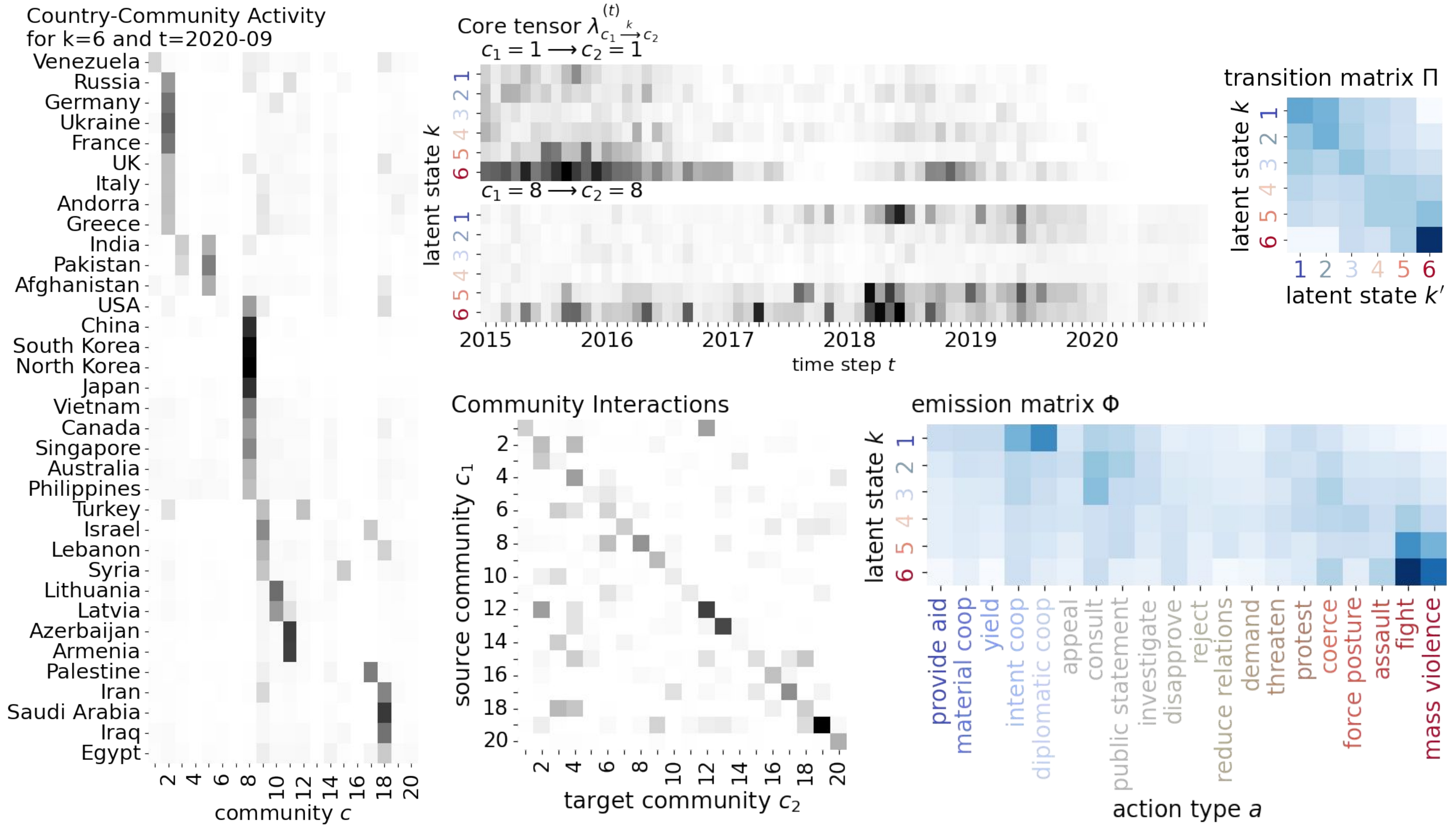}
 \caption{Posterior mean of parameters of \DPTM, with $K = 6$ latent states and $C=20$ latent communities, fitted to full temporal range (2015-2020) of ICEWS data. We find that the probability mass of the transition matrix is centered along the diagonal revealing step-wise (de-)escalatory dynamics. There is high probability of staying in state $k=6$ indicating that conflictual relationships may be hard to escape. The country-community affiliation matrix $\Psi$ provides no information on whether communities represent allies or enemies per se. To obtain this information, we interact the country-community matrix with the core tensor $\psi^{(\rightarrow)}_{c_1i} \sum_{c2=1}^C \sum_{j=1}^{\bV} \psi^{(\leftarrow)}_{c_2j} \lambda^{(t)}_{c_1 \xrightarrow{k} c_2}$ for specific choice of $k$ and $t$. 
 }
 \label{fig:real_results_15_20}
\end{figure*}

\begin{figure}[h]
  \centering
  \begin{minipage}[b]{0.48\textwidth}
     \includegraphics[width=1.0\linewidth]{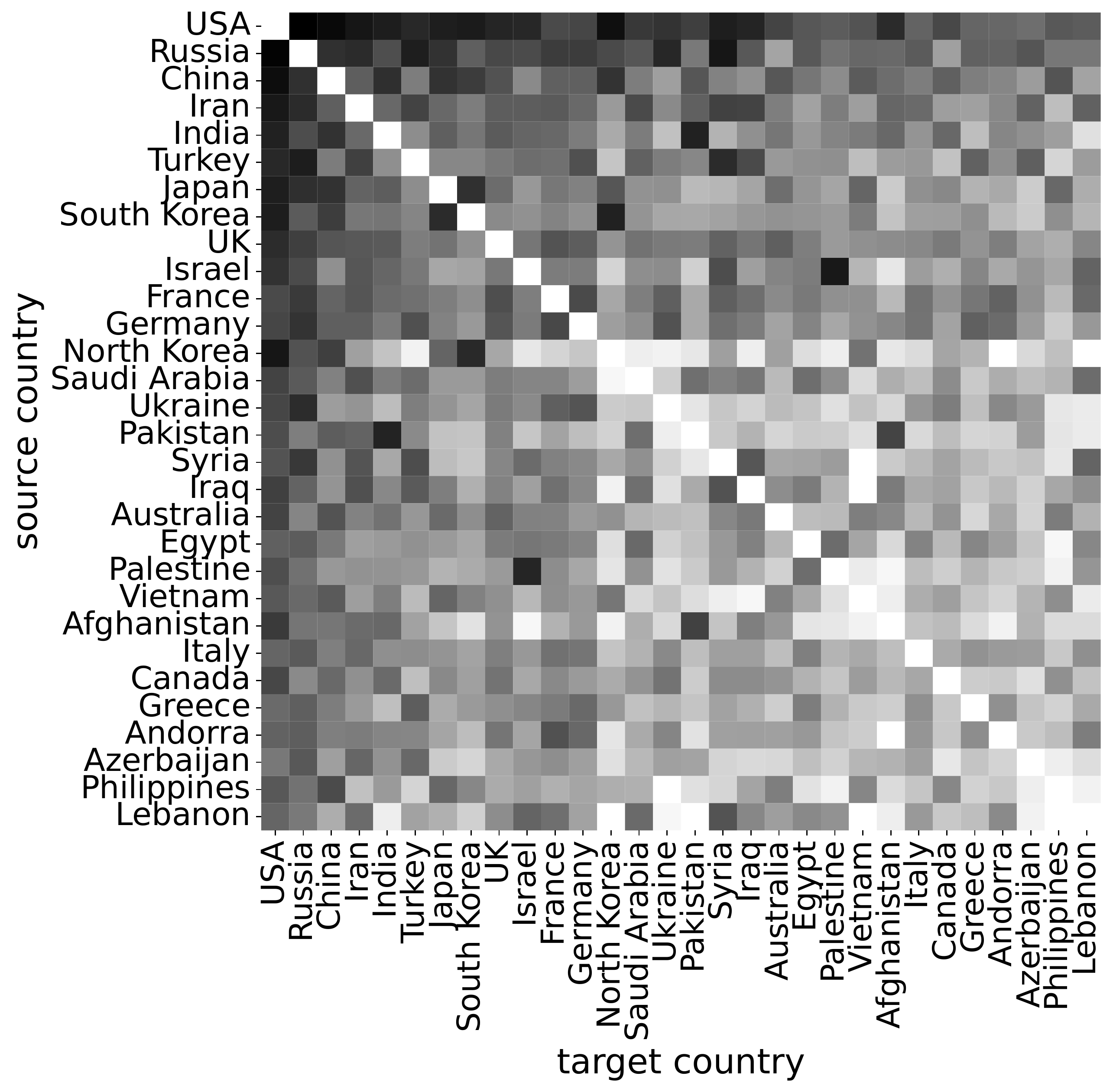} 
     \caption{Descriptive statistics showing total number of interactions between countries in ICEWS data from 2015 to 2020. The rows and columns are sorted by the total number of actions a country is involved in. Note that we omit self-targeted actions as indicated by the blank diagonal.
     }
     \label{fig:model}
  \end{minipage}
  \hfill
  \begin{minipage}[b]{0.48\textwidth}
     \centering
     \includegraphics[width=0.8\linewidth]{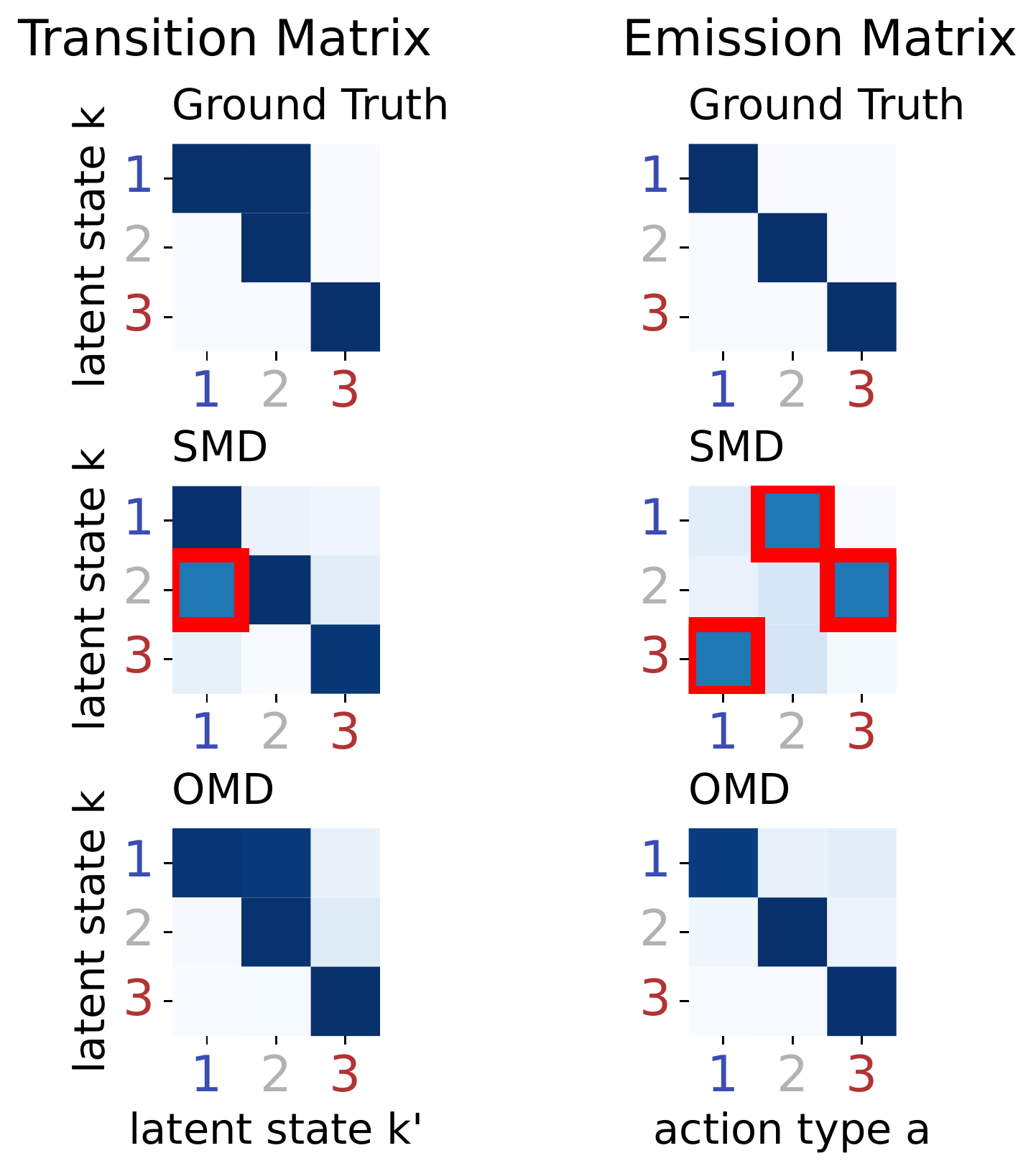} 
     \caption{Recovering ground truth structures in transition and emission matrices of a state-space model. Conventionally, rows are samples independently from a (standard) Dirichlet distribution. This can result in label switching making the latent states (topics) difficult to interpret. This is particularly problematic if states are ordinal, e.g., representing \setquote{ally}, \setquote{neutral} and \setquote{enemy} relations.}
     \label{fig:dir_label_switching}
  \end{minipage}
\end{figure}

 \begin{figure*}[h]
    \centering
    \includegraphics[width=0.9\linewidth]{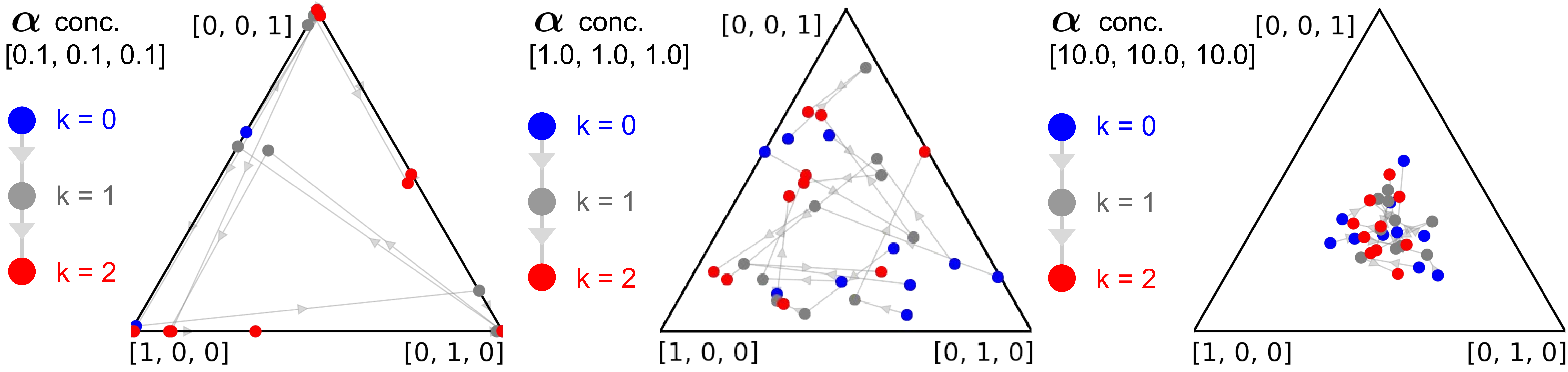} 
     \caption{Samples from the \SMD (\smd). Each point in the triangle plot represents a sample from a Dirichlet over $A=3$ classes. The $K=3$ points connected by a line represent an (unordered) sample from the \smd.}
     \label{fig:smd_samples}
\end{figure*}

\begin{figure}[h]
  \centering
  \begin{minipage}[b]{0.48\textwidth}
    \centering
    \includegraphics[width=1.0\linewidth]{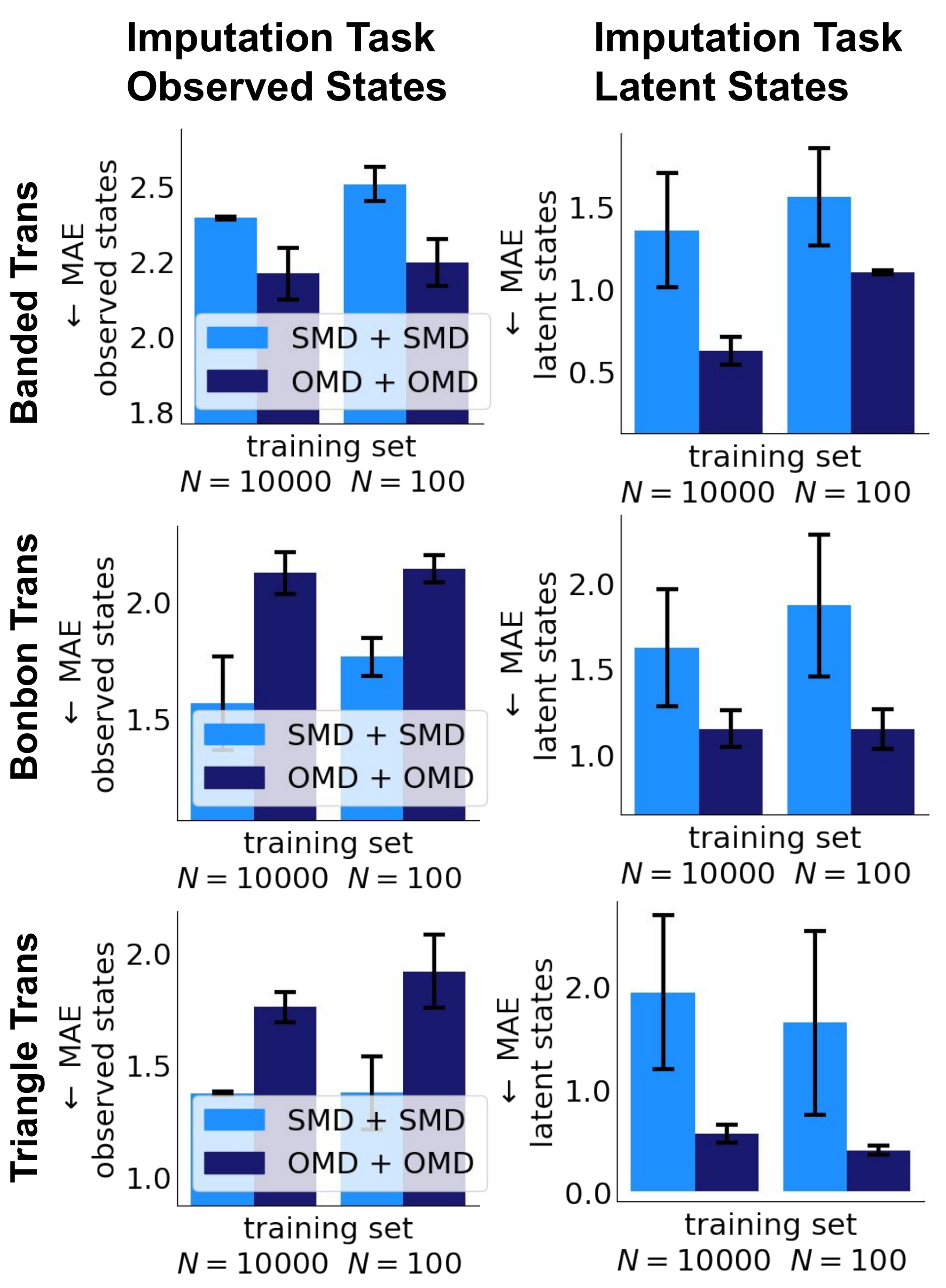}
    \caption{Imputation results of synthetic data experiments. As discussed in \cref{sec:synthetic_data}, we generate time series with different ground truth transition structures: \setquote{banded}, \setquote{bonbon}, \setquote{triangle}. We fit a Hidden Markov Model (HMM) to a train set of these data and evaluate imputation performance on a test set. In contrast to the forecasting experiments (\cref{fig:synth_forecast_hmm}), \smd + \smd outperforms \omd + \omd in two out of three cases on observed states. In contrast to forecasting, imputation does not necessarily require a model with temporal dynamics and the ordered transition matrix does not help. As expected, \omd + \omd performs better at imputing latent states because it circumvents label switching.}
    \label{fig:synth_impute_hmm}
  \end{minipage}
  \hfill
  \begin{minipage}[b]{0.48\textwidth}
\fontsize{10}{10}\selectfont
\centering
\renewcommand{\arraystretch}{1.5} 
\setlength{\tabcolsep}{0.1em} 
\begin{tabular}{clc}
\textbf{\begin{tabular}[c]{@{}c@{}}action type\\ $\textit{a}$\end{tabular}} & \multicolumn{1}{c}{\textbf{\begin{tabular}[c]{@{}c@{}}action\\ name\end{tabular}}} & \textbf{\begin{tabular}[c]{@{}c@{}}Goldstein\\ value\end{tabular}} \\
\multicolumn{1}{l}{}                                                        &                                                                                    & \multicolumn{1}{l}{}                                               \\
0                                                                           & provide aid                                                                        & 7.0                                                                \\
1                                                                           & engage material cooperation                                                        & 6.0                                                                \\
2                                                                           & yield                                                                              & 5.0                                                                \\
3                                                                           & express intent cooperate                                                           & 4.0                                                                \\
4                                                                           & engage diplomatic cooperation                                                      & 3.5                                                                \\
5                                                                           & appeal                                                                             & 3.0                                                                \\
6                                                                           & consult                                                                            & 1.0                                                                \\
7                                                                           & make public statement                                                              & 0.0                                                                \\
9                                                                           & investigate                                                                        & -2.0                                                               \\
10                                                                          & disapprove                                                                         & -2.0                                                               \\
11                                                                          & reject                                                                             & -4.0                                                               \\
12                                                                          & reduce relations                                                                   & -4.0                                                               \\
13                                                                          & demand                                                                             & -5.0                                                               \\
14                                                                          & threaten                                                                           & -6.0                                                               \\
15                                                                          & protest                                                                            & -6.5                                                               \\
16                                                                          & coerce                                                                             & -7.0                                                               \\
17                                                                          & exhibit force posture                                                              & -7.2                                                               \\
18                                                                          & assault                                                                            & -9.0                                                               \\
19                                                                          & fight                                                                              & -10.0                                                              \\
20                                                                          & unconventional mass violence                                                       & -10.0                                                             
\end{tabular}
\caption{Ordered \href{http://data.gdeltproject.org/documentation/CAMEO.Manual.1.1b3.pdf}{CAMEO action types} with assigned \href{https://parusanalytics.com/eventdata/cameo.dir/CAMEO.SCALE.txt}{Goldstein values}. We order action types by Goldstein value first and, in case of a tie, by CAMEO ID second.}
\label{tab:cameo_goldstein}
  \end{minipage}
\end{figure}

\end{document}